# Development of a Multi-Sensor Perceptual System for Mobile Robot and EKF-based Localization

T. T. Hoang, P. M. Duong, N. T. T. Van, D. A. Viet and T. Q. Vinh
Department of Electronics and Computer Engineering
University of Engineering and Technology
Vietnam National University, Hanoi

*Abstract*—**This paper presents the design and implementation of a perceptual system for the mobile robot using modern sensors and multi-point communication channels. The data extracted from the perceptual system is processed by a sensor fusion model to obtain meaningful information for the robot localization and control. Due to the uncertainties of acquiring data, an extended Kalman filter was applied to get optimal states of the system. Several experiments have been conducted and the results confirmed the functioning operation of the perceptual system and the efficiency of the Kalman filter approach.**

## I. INTRODUCTION

Based on advanced material technologies, modern sensors can be nowadays equipped for the mobile robot such as optical incremental encoders, heading sensors, ultrasonic sensors, infrared sensors, laser range finders and vision systems. These sensors combined with multi-point communication channels forms a perceptual system which allows the mobile robot to retrieve various parameters of the environment. Depending on the level of perception, the mobile robot has ability to perform certain tasks such as the localization, obstacle avoidance and path planning.

In a real application, sensors are often selected so as to accord with the goal of application, the specific constraints of the working environment, and the individual properties of the sensors themselves. Nevertheless, there is no single sensor which can adequately capture all relevant features of a real environment. It is necessary to combine the data from different sensors into a process known as *sensor fusion*. The expectation is that the fused data is more informative and synthetic than the original.

Several methods have been reported to cope with this trend. Durrant-Whyte has developed a multi-Bayesian estimation technique for combining touch and stereo sensing [1], [2]. Tang and Lee proposed a generic framework that employed a sensor-independent, feature-based relational model to represent information acquired by various sensors [3]. In [4], a Kalman filter update equation was developed to obtain the correspondence of a line segment to a model, and this correspondence was then used to correct position estimation. In [5], an extended Kalman filter was conducted to manipulate image and spatial uncertainties.

In this work, we develop a multi-sensor perceptual system for the mobile robot. Sensors include but not limit to optical quadratute encoders, compass sensors, ultrasonic sensors, laser range finders, global positioning systems (GPS) and vision systems. The goal is to equip the robot with diverse levels of perception to support a wide range of navigating applications including Internet-based telecontrol, semi-autonomy, and autonomy. At this stage of research, the optical quadratue encoders is used for position measurement, the compass sensor is used for deflect angle calculation and these data are fused inside an extended Kalman filter (EKF) to obtain optimal estimation of the robot position as well as reducing the uncertainties in measurements. Outputs of the EKF combined with the boundaries of objects detected from the LRF ensure the success navigation of the mobile robot in indoor environments.

The paper is arranged as follows. Details of the perceptual system are described in Section II. The algorithm for sensor fusion using EKF is explained in Section III. Section IV introduces the simulations and experiments. The paper concludes with an evaluation of the system, with respect to its strengths and weaknesses, and with suggestions of possible future developments.

## II. PERCEPTUAL SYSTEM DESIGN

The design of perceptual system is split into three perspectives: the communication, the sensor and the actuator; each is developed with the feasibility, the flexibility and the extendibility in mind. Fig.1 shows an overview of the system.

### A. Communication configuration

The communication is performed via low-rate and high-rate channels. The low-rate channel with standards of RS-485 is developed by an on-board 60MHz Microchip dsPIC30F4011-based micro-controller and MODBUS protocol for multi-point interface. The high-rate channels use the USB-to-COM and IEEE-1394/firewire available commercial ports.

### B. Sensors and Actuators

The sensor and actuator module contains basic components for motion control, sensing and navigation. These components are drive motors for moving control, sonar ranging sensors for obstacle avoidance, compass and GPS sensors for heading and global positioning, and laser range finder (LMS) and visual sensor (camera) for mapping and navigating.

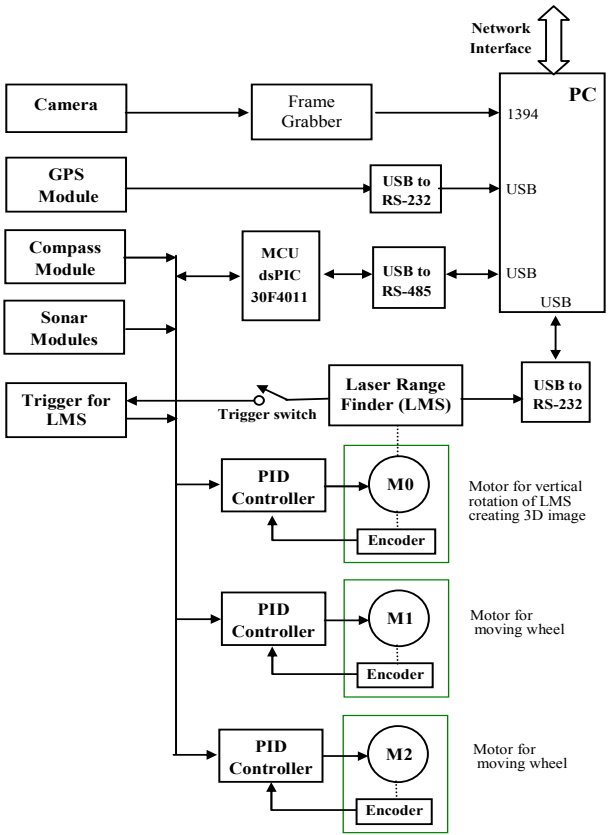

Figure 1. Sensors in relation with actuators and communication channels in the mobile robot

The drive system uses high-speed, high-torque, reversible-DC motors. Each motor is attached a quadrature optical shaft encoder that provides 500 ticks per revolution for precise positioning and speed sensing. The motor control is implemented in a microprocessor-based electronic circuit with an embedded firmware which permits to control the motor by a PID algorithm.

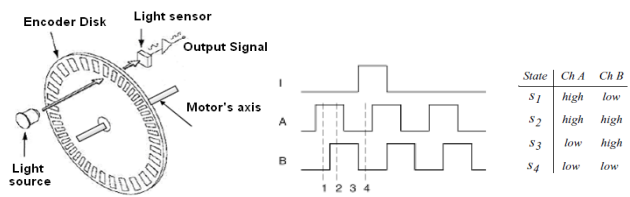

Figure 2. Optical encoder structure and output pulses

In the system, *optical quadrature encoders* are used. An optical encoder is basically a mechanical light chopper that produces a certain number of sine or square wave pulses for each shaft revolution. As the diameter of wheel and the gear coefficient are known, the angular position and speed of wheel can be calculated. In the quadrature encoder, a second light chopper is placed 90 degrees shifted with respect to the original resulting in twin square waves as shown in fig.2. Observed phase relationship between waves is employed to determine the direction of the rotation. In the system, measurements from encoders are used as input data for positioning and feedback for a closed-loop motor speed controller.

The heading sensor is used to determine the robot orientation. This sensory module contains a CMPS03 compass sensor operating based on Hall effect with heading resolution of 0.1° (fig.3). The module has two axes, x and y. Each axis reports the strength of the magnetic field component paralleled to it. The microcontroller connected to the module uses synchronous serial communication to get axis measurements [6].

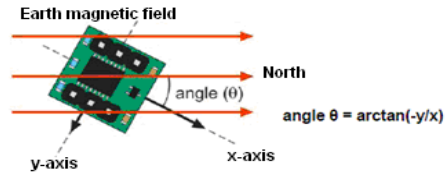

Figure 3. Heading module and output data

The GPS is mainly applied for positioning in the outdoor environment. A HOLUX GPS UB-93 module is used [7]. Due to the presence of networking in our system, an Assisted GPS (A-GPS) can be also used in order to locate and utilize satellite information of the network in the poor signal condition.

The system provides eight SFR-05 ultrasonic sensors split into four arrays, two on each, arranged at four sides of the robot. The measuring range is from 0.04m to 4m.

The vision system is detachable and mounted on the top of the robot. It mainly consists of an Omni-directional digital camera Hyper-Omni Vision SOIOS 55 with a high-rate IEEE-1394 data transfer line. With a 360-degree field of view, the Omni-directional camera is a good vision sensor for landmark-based navigation [8].

Last but not least, a 3D laser range finder (LRF) with a range from 0.04m to 80m is developed for the system. Its operation is based on the time-of-flight measurement principle. A single laser pulse is emitted out and reflected by an object surface within the range of the sensor. The lapsed time between emission and reception of the laser pulse is used to calculate the distance between the object and the sensor. By an integrated rotating mirror, the laser pulses sweep a radial range in its front so that a 2D field/area is defined.

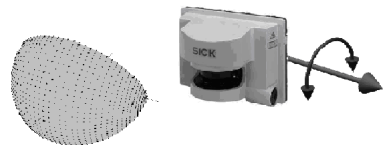

Figure 4. Three-dimension laser scanning plane

Due to the fixation of the pitching angle in the scanning plane, the information of 2D image may be, in certain cases, insufficient for obstacle detection. In those situations, a 3D image is necessary (fig.4). As the 2D scanner is popular and low-cost, building a 3D LRF from the 2D is usually a prior choice [9]. In our system, a 3D LRF is developed based on the 2D SICK-LMS 221 [9]. It has the view angle of 180° in horizontal and 25° in vertical. During the measurement time, a set of deflect angle $\beta$ and distance $R$ values are received. The set $(\beta, R)$ is then combined with the pitching angle $\alpha$ to create the 3D data. Based on these data, we can define the Cartesian coordinates of one point in the space.

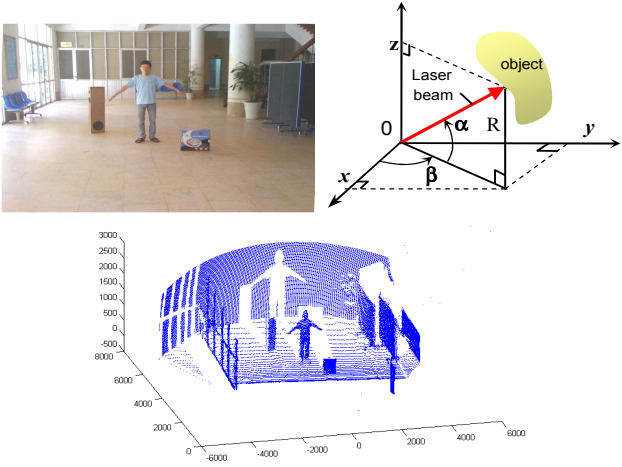

Figure 5.  A 3D-image captured by the LRF

Fig.6 shows the proposed sensor system implemented in the MSSR mobile robot developed at our laboratory.

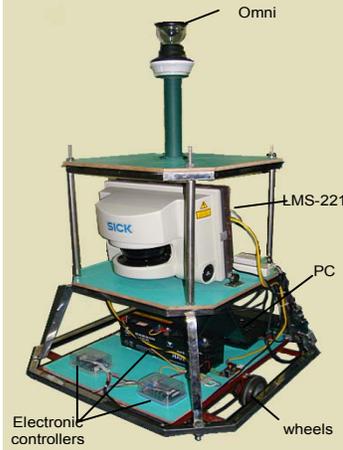

Figure 6.  The perceptual system in a mobile robot

### III. EKF-BASED LOCALIZATION

The developed sensor system enables the robot to perceive many parameters of the environment and consequently allows the robot to perform various tasks depending on requirements of the application. In this work, an investigation in the robot localization problem which is a fundamental condition for autonomous navigation is proposed. The aim is to determine the robot position during operation as accurately as possible.

The two wheeled, differential-drive mobile robot with non-slipping and pure rolling is considered. Fig.7 shows the coordinate systems and notations for the robot, where $(X, Y)$ is the global coordinate, $(X', Y')$ is the local coordinate relative to the robot chassis. $R$ denotes the radius of driven wheels, and $L$ denotes the distance between the wheels. The velocity vector $\mathbf{q} = [v\ \omega]^T$ consists of the translational velocity of the center of robot and the rotational velocity with respect to the center of robot. The velocity vector $\mathbf{q}$ and the posture vector $\mathbf{P_c} = [x_c\ y_c\ \theta_c]^T$ are associated with the robot kinematics as follows:

$$\dot{\mathbf{P}}_c = \begin{bmatrix} \dot{x}_c \\ \dot{y}_c \\ \dot{\theta}_c \end{bmatrix} = \begin{bmatrix} \cos\theta & 0 \\ \sin\theta & 0 \\ 0 & 1 \end{bmatrix} \begin{bmatrix} v \\ \omega \end{bmatrix} = \mathbf{J}(\theta)\mathbf{q} \qquad (1)$$

$$\mathbf{q} = \begin{bmatrix} v \\ \omega \end{bmatrix} = \begin{bmatrix} \frac{1}{2} & \frac{1}{2} \\ \frac{1}{L} & \frac{1}{L} \end{bmatrix} \begin{bmatrix} v_R \\ v_L \end{bmatrix} \qquad (2)$$

During one sampling period $\Delta t$, the rotational speed of the left and right wheels $\omega_L$ and $\omega_R$ create corresponding increment distances $\Delta s_L$ and $\Delta s_R$ traveled by the left and right wheels respectively:

$$\Delta s_L = \Delta t R \omega_L \qquad \Delta s_R = \Delta t R \omega_R \qquad (3)$$

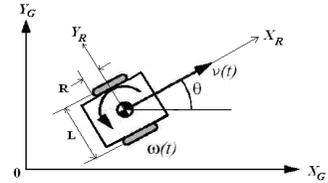

Figure 7.  The robot's pose and parameters

These can be translated to the linear incremental displacement of the robot center $\Delta s$ and the robot orientation $\Delta\theta$:

$$\Delta s = \frac{\Delta s_L + \Delta s_R}{2}$$
$$\Delta\theta = \frac{\Delta s_R - \Delta s_L}{L} \qquad (4)$$

The coordinates of the robot at time $k+1$ in the global coordinate frame can be then updated by:

$$\begin{bmatrix} x_{k+1} \\ y_{k+1} \\ \theta_{k+1} \end{bmatrix} = \begin{bmatrix} x_k \\ y_k \\ \theta_k \end{bmatrix} + \begin{bmatrix} \Delta s_k \cos(\theta_k + \Delta\theta_k/2) \\ \Delta s_k \sin(\theta_k + \Delta\theta_k/2) \\ \Delta\theta_k \end{bmatrix} \qquad (5)$$

In practice, (5) is not really accurate due to unavoidable errors appeared in the system. Errors can be both systematic such as the imperfectness of robot model and nonsystematic such as the slip of wheels. These errors have accumulative characteristic so that they can break the stability of the system if appropriate compensation is not considered. In our system, the compensation is carried out by the EKF.

Let $\mathbf{x} = [x\, y\, \theta]^T$ be the state vector. This state can be observed by some absolute measurements, $\mathbf{z}$. These measurements are described by a nonlinear function, $h$, of the robot coordinates and an independent Gaussian noise process, $\mathbf{v}$. Denoting the function (5) is $f$, with an input vector $\mathbf{u}$, the robot can be described by:

$$\mathbf{x}_{k+1} = f(\mathbf{x}_k, \mathbf{u}_k, \mathbf{w}_k)$$
$$\mathbf{z}_k = h(\mathbf{x}_k, \mathbf{v}_k) \quad (6)$$

where the random variables $\mathbf{w}_k$ and $\mathbf{v}_k$ represent the process and measurement noise respectively. They are assumed to be independent to each other, white, and with normal probability distributions: $\mathbf{w}_k \sim \mathbf{N}(0, \mathbf{Q}_k) \quad \mathbf{v}_k \sim \mathbf{N}(0, \mathbf{R}_k) \quad E(\mathbf{w}_i \mathbf{v}_j^T) = 0$

The steps to calculate the EKF are then realized as below:

1. Time update equations:

$$\hat{\mathbf{x}}_k^- = \mathbf{f}(\hat{\mathbf{x}}_{k-1}, \mathbf{u}_{k-1}, \mathbf{0}) \quad (7)$$

$$\mathbf{P}_k^- = \mathbf{A}_k \mathbf{P}_{k-1} \mathbf{A}_k^T \mathbf{W}_k \mathbf{Q}_{k-1} \mathbf{W}_k^T \quad (8)$$

2. Measurement update equations:

$$\mathbf{K}_k = \mathbf{P}_k^- \mathbf{H}_k^T (\mathbf{H}_k \mathbf{P}_k^- \mathbf{H}_k^T + \mathbf{V}_k \mathbf{R}_k \mathbf{V}_k^T)^{-1} \quad (9)$$

$$\hat{\mathbf{x}}_k = \hat{\mathbf{x}}_k^- + \mathbf{K}_k(\mathbf{z}_k - h(\mathbf{x}_k^-, \mathbf{0})) \quad (10)$$

$$\mathbf{P}_k = (\mathbf{I} - \mathbf{K}_k \mathbf{H}_k) \mathbf{P}_k^- \quad (11)$$

where: **P** is the covariance matrix of state variable $x$.

**Q** is the covariance of process disturbance $u$.

**K** is the Kalman gain.

**A** is the Jacobian matrix of partial derivates of **f** to $x$.

**W** is the Jacobian matrix of partial derivates of **f** to $w$.

**H** is the Jacobian matrix of partial derivates of **h** to $x$.

$\hat{\mathbf{x}}_k^- \in \Re^n$ is the *priori* state estimate at step $k$ given knowledge of the process prior to step $k$, and $\hat{\mathbf{x}}_k \in \Re^n$ is the *posteriori* state estimate at step $k$ given measurement $\mathbf{z}_k$.

From the above process, it is recognizable that the efficiency of EKF mainly depends on the estimation of white Gaussian noises $\mathbf{w}_k$ and $\mathbf{v}_k$. The noises $\mathbf{w}_k$ and $\mathbf{v}_k$ in its turn are featured by covariance matrices $\mathbf{Q}_k$ and $\mathbf{R}_k$ respectively.

$\mathbf{Q}_k$ is the input-noise covariance matrix depending on the standard deviations of noise of the right-wheel rotational speed and the left-wheel rotational speed. They are modeled as being proportional to the rotational speed $\omega_{R,k}$ and $\omega_{L,k}$ of these wheels at step $k$. This results in the variances equal to $\delta\omega_R^2$ and $\delta\omega_L^2$, where $\delta$ is a constant determined by experiments. The input-noise covariance matrix $\mathbf{Q}_k$ is defined as:

$$\mathbf{Q}_k = \begin{bmatrix} \delta\omega_{R,k}^2 & 0 \\ 0 & \delta\omega_{L,k}^2 \end{bmatrix} \quad (12)$$

$\mathbf{R}_k$ is a 3x3 diagonal matrix. The two elements $r_{11}$ and $r_{22}$ are extracted from the odometry data. To determine the element $r_{33}$ of $\mathbf{R}_k$, which is related to the deflect angular, the compass sensor is used. Let $\theta_k^C$ and $\theta_k^E$ are respectively the deflect angular of the robot determined by the compass sensor and the encoders at time $k$. $r_{33}$ is estimated by:

$$r_{33} = \frac{1}{N} \sum_{k-N}^{k} (\theta_k^C - \theta_k^E)^2 \quad (13)$$

IV. SIMULATIONS AND EXPERIMENTS

To evaluate the functioning operation of the sensor system and the EKF-based localization, several simulations and experiments have been conducted.

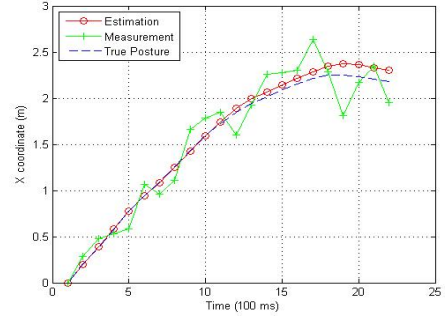

Figure 8. The true moving path of the robot and its estimations

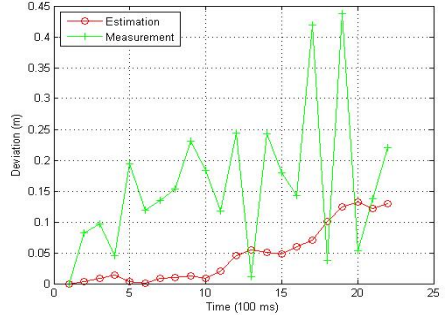

Figure 9. Deviation between the estimated paths and the true one

*A. Simulations*

Simulations are carried out in MATLAB in which the parameters are extracted from the real system. Fig.8 shows the true path of the robot in horizon direction and its estimations from the odometry and the EKF. Fig.9 shows the deviation between the estimated paths and the true path.

*B. Experiments*

Experiments are implemented in a rectangular shaped flat-wall environment constructed from several wooden plates surrounded by a cement wall. The robot is a two wheeled, differential-drive mobile robot. Its wheel diameter is *10 cm* and the distance between two drive wheels is *60cm*.

The drive motors are controlled by microprocessor-based electronic circuits. Due to the critical requirement of accurate speed control, the PID algorithm is implemented. The stability of motor speed checked by a measuring program written by LABVIEW is *±5%*. In case of straight moving, the speed of both wheels is set to *0.3m/s*. In turning, the speed of one wheel is reduced to *0.05m/s* in order to force the robot to turn to that wheel direction.

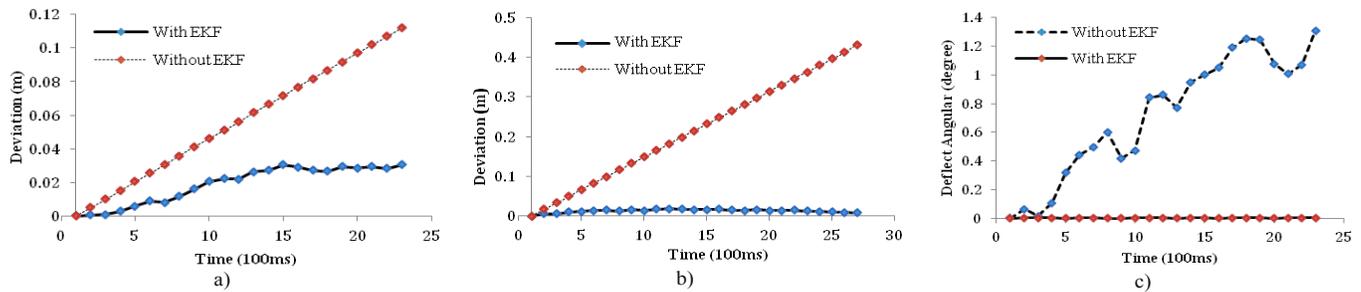

Figure 10. Deviation between the practical trajectories and true ones
a) Deviation in y coordinates     b) Deviation in x coordinates     c) Deviation in heading angles θ

The compass sensor has the accuracy of $0.1^0$. The sampling time *ΔT* of the EKF is *100ms*. The proportional factor $\delta$ of the input-noise covariance matrix $\mathbf{Q}_k$ is experimentally estimated as follows.

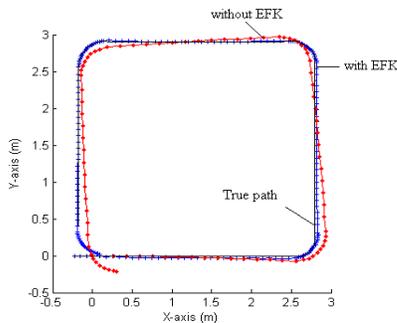

Figure 11. Trajectories of the robot moving along predefined paths with and without EKF

The differences between the true robot position and the position estimated by the kinematic model when driving the robot straight forwards several times (from the minimum to the maximum tangential speed of the robot) is observed. The differences between the true robot orientation and the orientation estimated by the kinematic model when rotating the robot around its own axis several times (from the minimum to the maximum angular speed of the robot) is also observed.

Based on the error in position and orientation, the parameter δ is calculated. In our system, the δ is estimated to be the value 0.01.

To evaluate the efficiency of sensor fusion using EKF, we programmed the robot to follow predefined paths under two different configurations: with and without the EKF. Fig.11 shows the trajectories of the robot moving along a rounded rectangular path in which the one with dots corresponds to the non-existence of EKF in configuration and the one with pluses corresponds to the existence of EKF. Fig.10 shows the deviation between the practical trajectories and true ones. Different paths such as rounded rectangular and arbitrary curves are also experimented and it is concluded from the results that the fusion algorithm significantly improves the robot localization.

V. CONCLUSION

A perceptual system for the mobile robot was developed with many sensors including position speed encoders, integrated sonar ranging sensors, compass and laser finder sensors, the global positioning system (GPS) and the visual system. An EKF was designed to fuse the raw data of sensors. Simulations and experiments show that this combination approach significantly improves the accuracy of robot localization and is sufficient to ensure the success of robot navigation. Further investigation will be continued with more sensor combination to better support the localization in outdoor environments.

ACKNOWLEDGMENT

This work was supported by TRIGB Project, University of Engineering and Technology, Vietnam National University, Hanoi.

REFERENCES

[1] H. F. Dunant-Whyte, "Consistent integration and propagation of disparate sensor observations," Znt. J. Robot. Res., vol. 6, no. 3, pp. 3-24, 1987.

[2] H. F. Dunant-Whyte , "Sensor models and multi-sensor integration,"Z nt.J.Robot. Res., vol. 7, no. 6, pp. 97-113, 1988.

[3] Y. C. Tang and C. S. G. Lee, "A geometric feature relation graph formulation for consistent sensor fusion," in Proc. IEEE 1990 Int. ConSyst., Man, Cybern., Los Angeles, CA, 1990, pp. 188-193.

[4] J. L. Crowly, "World modeling and position estimation for a mobile robot using ultrasonic ranging," in Proc. IEEE Int. Conf. Robot., Automat., 1989, pp. 674-680.

[5] T. Skordas, P. Puget, R. Zigmann, and N. Ayache, "Building 3-D edge-lines tracked in an image sequence," in Proc. Intell. Autonomous Systems-2, Amsterdam, 1989, pp. 907-917.

[6] [Online] http://www.robot-electronics.co.uk/htm/cmps3tech.htm

[7] [Online] http://www.holux.com

[8] N. Winters et al, "Omni-directional vision for robot navigation", Omnidirectional Vision, 2000. Proceedings. IEEE Workshop on, Hilton Head Island , USA, Jun 2000.

[9] Sick AG., 2006-08-01 Telegrams for Operating/ Configuring the LMS 2xx (Firmware Version V2.30/X1.27), www.sick.com , Germany.